\documentclass{article}

\usepackage{PRIMEarxiv}

\usepackage[utf8]{inputenc} % allow utf-8 input
\usepackage[T1]{fontenc}    % use 8-bit T1 fonts
\usepackage{hyperref}       % hyperlinks
\usepackage{url}            % simple URL typesetting
\usepackage{booktabs}       % professional-quality tables
\usepackage{amsfonts}       % blackboard math symbols
\usepackage{nicefrac}       % compact symbols for 1/2, etc.
\usepackage{microtype}      % microtypography
\usepackage{lipsum}
\usepackage{fancyhdr}       % header
\usepackage{graphicx}       % graphics
\graphicspath{{media/}}     % organize your images and other figures under media/ folder
\usepackage{amsmath}
\usepackage{subcaption}     % Horizontal double image layout

\usepackage{caption} 
\usepackage{multirow}

\title{Information Extraction from Electricity Invoices with General-Purpose Large Language Models}

\author{
  Javier Gómez \\
  Centro de Tecnologías de la Imagen (CTIM)\\
  Instituto Universitario de Cibernética, Empresas y Sociedad (IUCES)\\
  Universidad de Las Palmas de Gran Canaria \\
  35017 Las Palmas de Gran Canaria, Spain\\
  \texttt{javier.gomez105@alu.ulpgc.es} \\
  \And
  Javier Sánchez \\
  Centro de Tecnologías de la Imagen (CTIM)\\
  Instituto Universitario de Cibernética, Empresas y Sociedad (IUCES)\\
  University of Las Palmas de Gran Canaria \\
  35017 Las Palmas de Gran Canaria, Spain\\
  \texttt{jsanchez@ulpgc.es} \\
}

\begin{document}
\maketitle

\begin{abstract}
Information extraction from semi-structured business documents remains a critical challenge for enterprise management. This study evaluates the capability of general-purpose Large Language Models to extract structured information from Spanish electricity invoices without task-specific fine-tuning. Using a subset of the IDSEM dataset, we benchmark two architecturally distinct models, Gemini 1.5 Pro and Mistral-small, across 19 parameter configurations and 6 prompting strategies. Our experimental framework treats prompt engineering as the primary experimental variable, comparing zero-shot baselines against increasingly sophisticated few-shot approaches and iterative extraction strategies. Results demonstrate that prompt quality dominates over hyperparameter tuning: the F1-score variation across all parameter configurations is marginal, while the gap between zero-shot and the best few-shot strategy exceeds 19 percentage points. The best configuration (few-shot with cross-validation) achieves an F1-score of 97.61\% for Gemini and 96.11\% for Mistral-small, with document template structure emerging as the primary determinant of extraction difficulty. These findings establish that prompt design is the critical lever for maximizing extraction fidelity in LLM-based document processing, thereby providing an empirical framework for integrating general-purpose LLMs into business document automation.
\end{abstract}


\section{Introduction}
The transition towards a data-driven economy has elevated document processing from an administrative task to a strategic capability. Business documents such as invoices, contracts, and purchase orders contain structured information essential for enterprise resource planning, financial forecasting, and operational decision-making. With an estimated 550 billion invoices generated each year~\cite{Koch2019einvoicing}, the automated extraction of information from these documents presents a significant challenge and an opportunity for operational transformation. Manual document processing is not only slow and expensive, but also prone to errors that propagate through enterprise systems.

Traditional approaches to document information extraction have evolved through several technological generations. Optical Character Recognition (OCR) systems enabled text detection but remained sensitive to image quality and typographic variation. Template-based systems improved extraction precision for standardized formats but failed when encountering layout variations. Some works~\cite{Barchard2011humanerror} have shown that manual data entry processes can be up to 2,958\% more error-prone than fully automated alternatives, underscoring the critical need for robust automation. Classical Machine learning approaches, such as Support Vector Machines (SVMs), Random Forest, or Decision Trees, introduced adaptability~\cite{Sanchez2024bagofwords}, but required extensive manually-annotated datasets for each document type, making scalability economically prohibitive.

The emergence of Large Language Models (LLMs) presents a paradigm shift in document processing. The field has experienced exponential growth since the release of ChatGPT in late 2022~\cite{Zhao2023llmsurvey}. These models, built upon the Transformer architecture~\cite{Vaswani2017attention}, exhibit remarkable capabilities for understanding context and semantic relationships through their self-attention mechanisms. Unlike template-based predecessors, LLMs can interpret documents based on meaning rather than spatial position, offering the potential for flexible, generalizable extraction without task-specific fine-tuning.

The high accessibility to these capabilities has transformed the landscape from model construction to model orchestration~\cite{Zhao2023llmsurvey}. Prompt engineering has emerged as a critical discipline for optimizing model behavior. \textit{Zero-shot} and \textit{few-shot} learning paradigms enable models to perform extraction tasks without extensive training data~\cite{schulhoff2025promptreportsystematicsurvey}. Recent work has demonstrated that providing LLMs with structured text formats, such as \textit{Markdown}, significantly improves their comprehension of document structure and hierarchical relationships~\cite{Braun2025hiddenstructure}.

However, the practical deployment of LLMs for high-fidelity information extraction raises fundamental questions that remain inadequately addressed. While academic benchmarks such as DocVQA~\cite{Mathew2021docvqa}, ChartQA~\cite{Masry2022chartqa}, and InfographicVQA~\cite{Mathew2022infographicvqa} measure general document understanding capabilities, there is a lack of rigorous empirical studies on specific business document types. The field can benefit from systematic investigation regarding which factors most significantly influence extraction quality based on model architecture, inference parameters~\cite{Renze2024temperature, Brown2020fewshot}, or instruction formulation on business documents.

This work aims to create a comprehensive empirical study evaluating LLM performance in extracting structured information from Spanish electricity invoices. Our investigation is structured around three core research questions: i) What extraction precision can general-purpose LLMs achieve without fine-tuning; ii) what is the relative impact of inference parameters versus prompt engineering strategies on extraction quality; iii) how do different document template structures influence model performance.

We employ a subset of the Invoices Database of the Spanish Electricity Market (IDSEM) dataset~\cite{Sanchez2022idsem}, comprising 75,000 invoices across 6 templates with 107 semantic labels, providing a controlled and realistic test set. This dataset was synthetically generated, using real data and statistics issued by public organizations and electricity companies. Our experimental framework relies on two different LLM models: Gemini 1.5 Pro~\cite{Reid2024gemini}, a large-scale Mixture-of-Experts model~\cite{Shazeer2017moe, Fedus2021switch}, and Mistral-small~\cite{Jiang2023mistral}, an efficient dense architecture employing Sliding Window Attention and Grouped-Query Attention optimizations for faster inference.

This study offers three primary contributions to the understanding of LLMs in information extraction. First, we demonstrate that prompt structure serves as a significantly more powerful lever for performance than hyperparameter tuning, which yielded only marginal gains in our tests. We establish that general-purpose LLMs, with appropriately designed prompts, can achieve high-fidelity extractions without fine-tuning. Finally, we identify document template structure as the fundamental determinant of extraction difficulty, showing that structural complexity in the source material is the main determinant of performance variance.

Section~\ref{se:relate_work} reviews related work in document understanding and LLM-based information extraction. Section~\ref{se:methods} describes the IDSEM dataset and how we processed it in our pipeline. Section ~\ref{se:methodology} describes the structure of the prompting methods. Section ~\ref{se:experiments} details the experimental methodology and evaluation metrics, and Section ~\ref{se:results} reports quantitative results. Section~\ref{se:conclusion} discusses the implications of our findings, limitations, and directions for future research.

%-------------------------- Related Work ---------------------------
\section{Related Work}
\label{se:relate_work}
Information extraction from business documents has been a persistent challenge in enterprise automation. The traditional approach relied on Optical Character Recognition (OCR) to convert document images into machine-readable text, followed by rule-based systems or template matching to locate specific fields. These are effective for highly standardized documents, but they fail when encountering layout variations, a common scenario in which each supplier generates invoices with varying formatting.

The field evolved through several generations of machine learning approaches. Traditional techniques, including logistic regression, Support Vector Machines (SVM), and Random Forest, were applied to information extraction tasks~\cite{Sanchez2024bagofwords}. However, these methods required extensive manually-annotated datasets for each document type, creating scalability bottlenecks that are economically prohibitive for most organizations when adapting to new formats. Document Layout Analysis (DLA) emerged to provide structural understanding, identifying and categorizing logical regions such as text blocks, tables, and headers before extraction. Deep learning revolutionized this task by framing DLA as object detection, with Transformer-based architectures like the Document Image Transformer (DIT)~\cite{DiT} learning spatial and structural relationships between document components.

The work in~\cite{Sanchez2024bagofwords} evaluated classical machine learning methods on the IDSEM dataset. Their approach converted PDF invoices to text, generated eleven-word sentences using a sliding window, and classified the central word using TF-IDF features combined with custom-designed features for electricity data. They achieved the highest global precision (91.86\%) with SVM and Radial Basis Functions for templates seen during training. However, precision dropped substantially to 67.20\% for unseen templates, revealing significant overfitting to training layouts. The authors explicitly identified Large Language Models as a direction for future work.

Large Language Models (LLMs) marked a paradigm shift in document processing, with the field experiencing exponential growth following the release of ChatGPT in late 2022~\cite{Zhao2023llmsurvey}. These models, built upon the Transformer architecture~\cite{Vaswani2017attention}, exhibit remarkable capabilities for understanding context and semantic relationships through self-attention mechanisms. The Mixture-of-Experts (MoE) architecture~\cite{Shazeer2017moe} increased model capacity without proportional computational cost by activating only a subset of expert networks for each token, with Switch Transformers demonstrating trillion-parameter models through efficient sparsity~\cite{Fedus2021switch}. While the prohibitive cost of full fine-tuning drove development of Parameter-Efficient Fine-Tuning methods such as LoRA~\cite{Hu2021lora} and QLoRA~\cite{Dettmers2023qlora}, prompt-based approaches have emerged as an alternative that requires no fine-tuning at all.

The most transformative evolution in Intelligent Document Processing (IDP) has been the shift from sequential pipelines to unified multimodal architectures. Traditional approaches where OCR, layout analysis, and natural language processing operated sequentially were fragile, with errors propagating without correction. LayoutLM~\cite{Xu2019layoutlm} introduced 2D position embeddings and image embeddings, enabling models to jointly learn language and layout while dramatically outperforming text-only approaches. Current multimodal LLMs like Gemini~\cite{Reid2024gemini} and open-source models such as Donut and LayoutLLM process document images directly to generate structured output, eliminating explicit OCR steps~\cite{Hong2024layoutllm}.

Perhaps the most impactful recent trend is the shift toward prompt-based extraction. Zero-shot learning enables models to process never-before-seen document types without specific training examples, leveraging vast language knowledge to infer structure and semantics~\cite{kojima2023largelanguagemodelszeroshot}, while few-shot learning achieves high extraction accuracy with only a small number of labeled examples~\cite{Gali2025fewshot}. Chain-of-thought prompting~\cite{Wei2022chainofthought} has demonstrated that decomposing complex tasks into sequential reasoning steps improves model performance on challenging problems. Recent work has shown that providing LLMs with structured text formats, such as Markdown, significantly improves their comprehension of document structure and hierarchical relationships~\cite{Braun2025hiddenstructure}.

Evaluation benchmarks such as DocVQA~\cite{Mathew2021docvqa}, ChartQA~\cite{Masry2022chartqa}, and InfographicVQA~\cite{Mathew2022infographicvqa} measure general document understanding capabilities, though gaps remain between academic benchmarks and real-world performance~\cite{Boiangiu2021due, Li2024vrdu}. The generation of factually incorrect content (hallucination) remains a critical challenge for LLM deployment~\cite{Ji2025hallucination}.

Despite these advances, rigorous empirical studies comparing LLM-based approaches against classical machine learning baselines on identical datasets remain scarce. Our work addresses this gap by evaluating general-purpose LLMs on the same IDSEM corpus. Furthermore, while inference parameters such as temperature, Top-p, and Top-k are known to affect generation behavior~\cite{Renze2024temperature}, their relative impact compared to prompt engineering strategies has not been quantified for high-fidelity extraction tasks. This work provides a large-scale empirical evaluation addressing both questions: whether LLMs can surpass classical ML performance without fine-tuning, and which factors most significantly influence extraction quality.

%-------------------------- Materials and Methods ---------------------------
\section{Dataset and preprocessing}
\label{se:methods}

This section presents the dataset used in this study, the preprocessing steps for LLM compatibility, and the selected subset. We detail the rationale for choosing the IDSEM corpus, describe the challenges encountered during adaptation, and explain the ground-truth refinement process necessary for evaluation.

\subsection{The IDSEM Dataset}

The development and evaluation of information extraction methods depend critically on the availability of large-scale, well-annotated corpora. In this work, we selected the IDSEM dataset, which is publicly available and comprises 75,000 synthetically generated invoices in PDF format. The dataset resolves the fundamental obstacle of data privacy that prevents access to real commercial documents by generating a synthetic corpus that is realistic in structure, content, and visual layout.

The IDSEM generation pipeline creates invoices indistinguishable from authentic documents. A simulation module generates field values by sampling from statistical distributions derived from official Spanish regulatory bodies. Then, an automated population system fills document templates in DOCX format, replacing placeholder codes (e.g., \texttt{\{\{A1\}\}}) with simulated data, and the documents are converted to PDF. Templates were created from real invoices of major Spanish marketers, including Iberdrola, Endesa, Naturgy, EDP, and Repsol.

\begin{figure}[htbp]
    \centering
    \begin{subfigure}[b]{0.48\textwidth}
        \centering
        \fbox{\includegraphics[width=\linewidth]{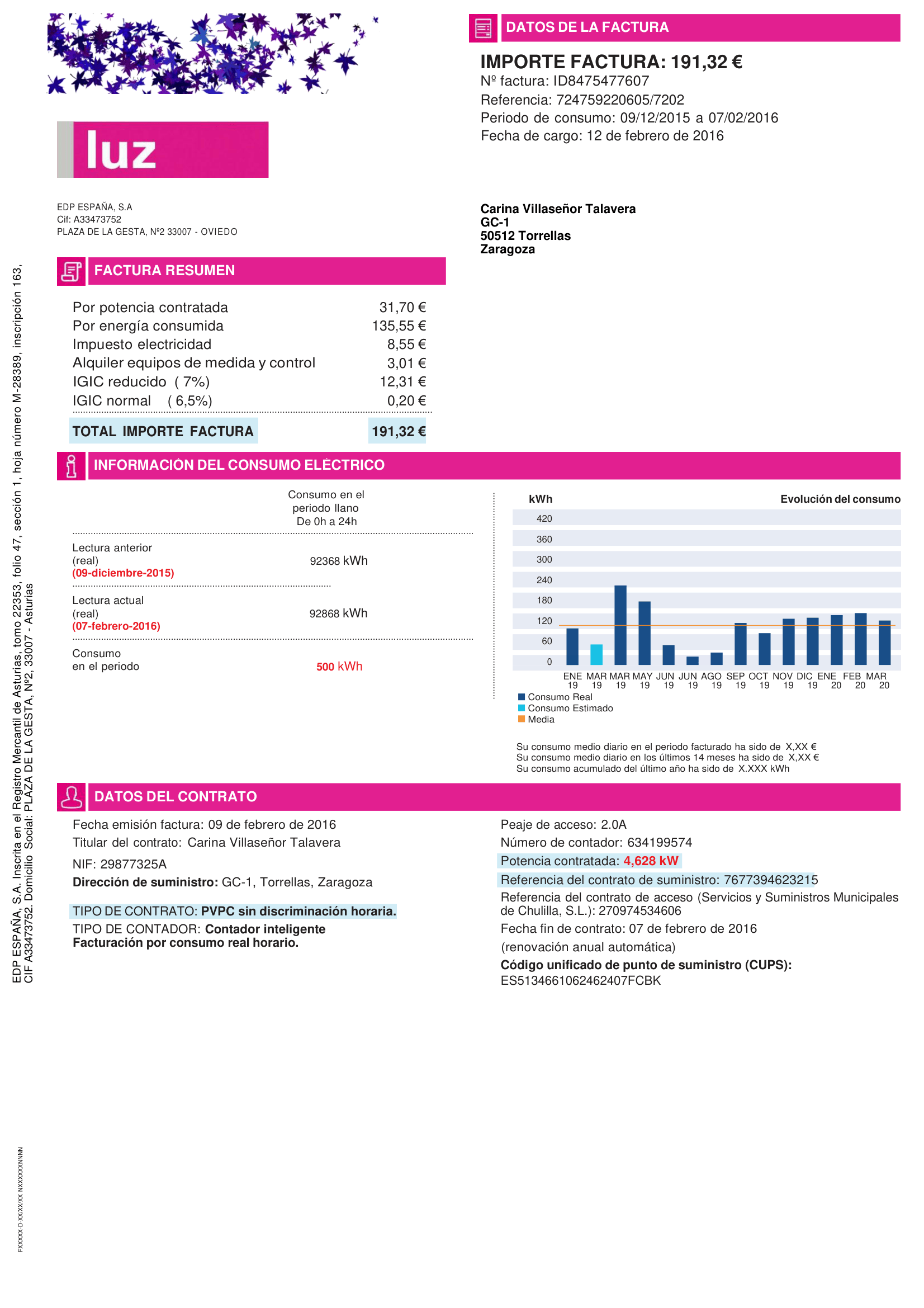}}
        \label{fig:invoice1}
    \end{subfigure}
    \hfill 
    \begin{subfigure}[b]{0.48\textwidth}
        \centering
        \fbox{\includegraphics[width=\linewidth]{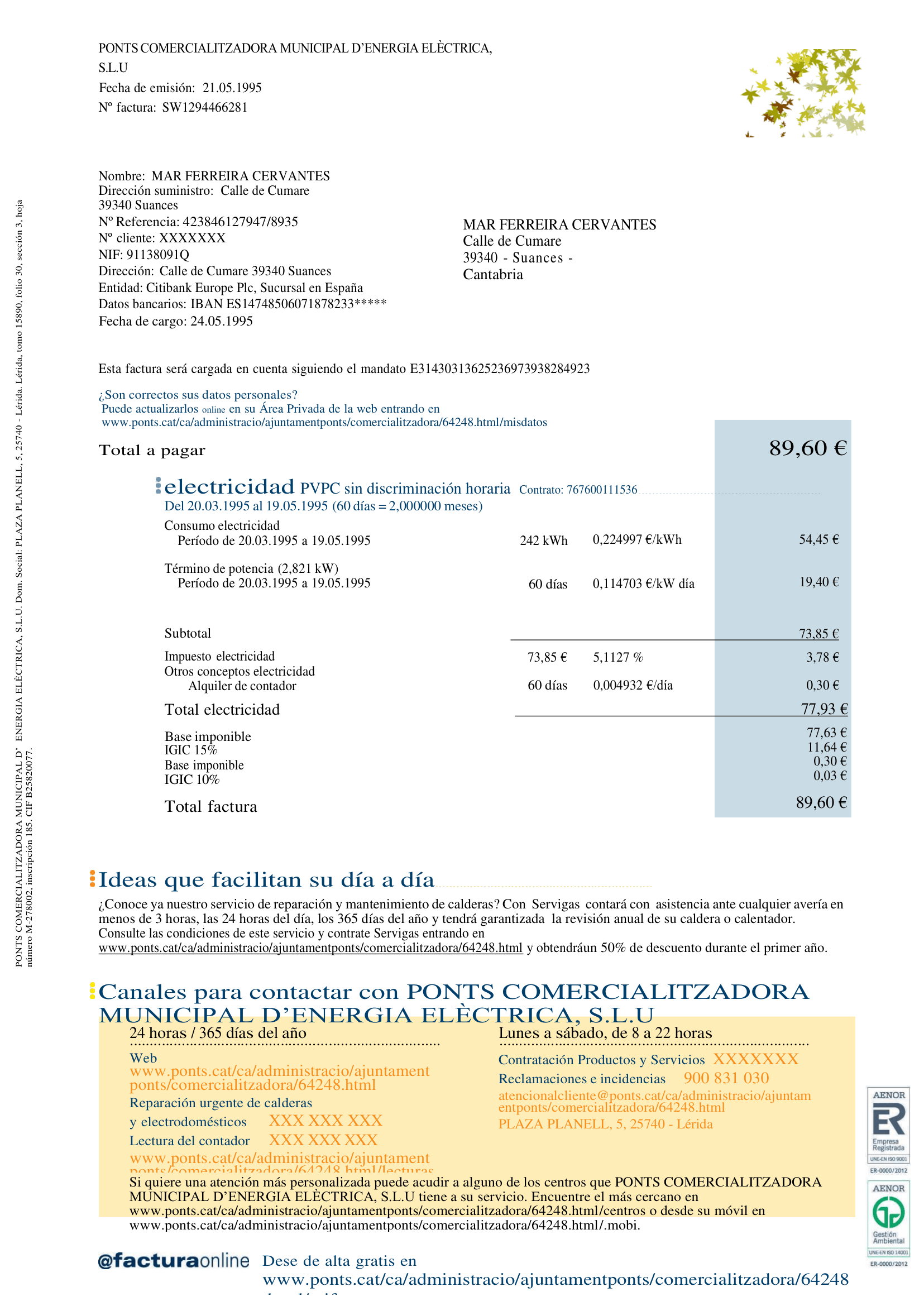}}
        \label{fig:invoice2}
    \end{subfigure}
    \caption{Examples of electricity invoices from the IDSEM dataset. Despite containing similar content, each template organizes information differently using tables, boxes, and columns. Source: Sánchez et al.~\cite{Sanchez2022idsem}.}
    \label{fig:invoices}
\end{figure}

The dataset is organized into a training directory of 30,000 documents across 6 templates, and a test directory containing 45,000 documents across 9 templates. Figure~\ref{fig:invoices} illustrates the visual diversity of invoice layouts across different marketers, demonstrating how identical semantic content is presented with varying structures, table formats, and spatial arrangements. Each training invoice is accompanied by a JSON file containing the ground-truth labels. Notably, three templates in the test set are not included in the training set, enabling evaluation of generalization to unseen document layouts.

The IDSEM annotation schema defines 86 distinct semantic concepts organized into 12 thematic categories, each identified by a letter code. Table~\ref{tab:label_categories} summarizes this organization. The first character of each label code indicates its category: for instance, \texttt{A1} represents the customer name (category A), while \texttt{J5} denotes the total invoice amount (category J). This hierarchical structure reflects the logical organization of electricity invoices.

\begin{table}[ht!]
\centering
\caption{IDSEM label categories and their semantic content. Each category groups related fields that appear together in invoice documents.}
\label{tab:label_categories}
\begin{tabular}{clc}
\hline
\textbf{ID} & \textbf{Description} & \textbf{Labels} \\
\hline
A & Customer receiving the invoice & 6 \\
B & Customer data as stated in contract & 6 \\
C & Marketer (retailer) information & 14 \\
D & Distributor information & 5 \\
E & Contract details & 9 \\
F & General invoice information & 8 \\
G & Customer financial information & 10 \\
I & Energy consumption data & 3 \\
J & Invoice breakdown summary & 5 \\
K & Detailed invoice breakdown & 10 \\
M & Equipment rental charges & 2 \\
N & Taxes and fees & 8 \\
\hline
\multicolumn{2}{r}{\textbf{Total semantic concepts}} & \textbf{86} \\
\hline
\end{tabular}
\end{table}

While the dataset documents 86 fundamental semantic concepts, the total number of extractable labels reaches 107 due to format variants. Twenty-one sub-labels represent alternative representations of the same field, primarily dates and monetary values. For example, a billing start date may appear in long format (\texttt{F4l}: ``12 de enero de 2021''), short format (\texttt{F4s}: ``12/01/2021''), or punctuated format (\texttt{F4p}: ``12.01.2021''). Similarly, rate values may be expressed annually (\texttt{K2}: €/kW per year) or daily (\texttt{K2d}: €/kW per day). This format multiplicity reflects real-world invoice heterogeneity and presents an additional challenge for extraction systems.

To enable LLM-based extraction, we developed a preprocessing pipeline that converts PDF invoices to structured text representations. The choice of Markdown as the target format is grounded in recent research~\cite{Braun2025hiddenstructure} demonstrating that structured text formats significantly improve LLM comprehension of document layout and hierarchical relationships. Unlike raw text extraction, Markdown preserves semantic structure through headers, tables, and formatting markers that align with how documents organize information visually.

During preprocessing, we encountered several challenges: First, the conversion tool could not extract text located in document margins, embedded in images, or contained within graphical elements, resulting in information loss for certain fields; second, the JSON ground truth files contain keys for all 107 possible labels, whereas each invoice template displays only a subset of these fields; third, the synthetic nature of the dataset meant some fields were annotated as present but contained empty values, creating evaluation ambiguity.

To address these issues, we created an explicit mapping documenting, for each of the six training templates, which subset of the 107 labels is actually present and extractable in the final Markdown representation. This registry proved indispensable during evaluation, ensuring that models were only penalized for fields genuinely visible in the input document rather than for labels that never appeared in that template. Table~\ref{tab:template_labels} shows the number of extractable labels per template after this filtering process.

\begin{table}[htbp]
\centering
\caption{Number of true labels per template after ground-truth refinement.}
\label{tab:template_labels}
\renewcommand{\arraystretch}{1.2}
\begin{tabular}{lcccccc}
\hline
\textbf{Template} & T1 & T2 & T3 & T4 & T5 & T6 \\
\hline
Extractable labels & 75 & 62 & 53 & 68 & 51 & 52 \\
\hline
\end{tabular}
\end{table}

We selected a subset of the training set for evaluation. From each template, 400 random invoices were obtained, yielding a total of 2,400 documents. This subset preserves the structural diversity of the full dataset while enabling comprehensive experimentation within practical resource constraints. The selection criteria ensured equal coverage of all template variants, preventing bias toward any particular document layout. %With approximately 60 extractable labels per invoice on average, the 2,400 document subset contains a substantial number of individual label extractions per experimental configuration, providing sufficient statistical power to detect meaningful performance differences across conditions.

All experiments were conducted on the training partition rather than the test set, as the latter lacks publicly available ground-truth labels. This approach aligns with prior work on the same dataset \cite{Sanchez2024bagofwords}, enabling direct comparison of results. The consistent use of the training partition across studies ensures that performance differences reflect methodological advances rather than evaluation set variations.

%-------------------------- Experiments ---------------------------
\section{Prompt Engineering for Document Extraction}
\label{se:methodology}

This section describes the prompt engineering strategies developed to adapt general-purpose LLMs to the task of structured information extraction from electricity invoices. Unlike fine-tuning approaches that modify model weights, prompt engineering operates entirely at inference time, crafting input sequences that guide the model toward desired behaviors. We designed six distinct strategies spanning zero-shot, few-shot, and iterative paradigms to evaluate their influence on extraction quality.

All strategies share a common architectural framework consisting of a system prompt and a user prompt. The system prompt establishes the model's role and behavioral constraints, instructing it to act as a specialized document extraction assistant that outputs the result in JSON format. %This role assignment leverages the model's instruction-following capabilities while constraining the output structure to streamline automated evaluation. It is important to note that this structured output specification defined entirely through natural language instructions is feasible only because the models employed are state-of-the-art foundational models with strong instruction-following capabilities. 
These models can produce consistent JSON outputs without being explicitly fine-tuned for structured generation; smaller or less capable models would likely require task-specific fine-tuning to achieve comparable output consistency.

The user prompt contains three elements: i) task instructions specifying the extraction objective; ii) the document content in Markdown format; and iii) the expected output schema listing all target fields with their semantic descriptions. The choice of Markdown as the document representation format is deliberate, as structured text formats preserve semantic relationships and visual hierarchies that help LLM understand the document; see Section~\ref{se:methods}.

The output schema provides the model with explicit field definitions, reducing ambiguity in label semantics. For instance, the schema distinguishes between \texttt{A1} (customer name on the invoice letter) and \texttt{B1} (customer name as stated in the client-company contract), which may differ in the same document. Each field definition includes a brief description, expected data type, and examples where applicable. This approach ensures a consistent output across all extractions, facilitating automated comparison against ground-truth data. 

We evaluated six prompting strategies varying along two dimensions: the presence of in-context examples (zero-shot vs. few-shot) and extraction granularity (holistic vs. iterative). Table~\ref{tab:prompting_strategies} describes each strategy in detail.

\begin{table}[htbp]
\centering
\caption{Summary of prompting strategies evaluated.}
\label{tab:prompting_strategies}
\renewcommand{\arraystretch}{1.3}
\begin{tabular}{llp{7.5cm}}
\hline
\textbf{Strategy} & \textbf{Paradigm} & \textbf{Description} \\
\hline
Zero-shot & Zero-shot & No examples; relies solely on task instructions and field definitions to measure inherent LLM extraction capability. \\
Few-shot\_v1 & Few-shot & One annotated example with explicit label markers (\texttt{\#A1 text \#A1}) signaling text-to-label correspondences. \\
Few-shot\_v2 & Few-shot & One non-annotated example (plain text), testing whether the model can infer correspondences from the input-output pair alone. \\
Cross-valid\_v1 & Few-shot & Three examples from templates T1, T2, T3. and then tested on templates T4, T5, T6 (sequential grouping). \\
Cross-valid\_v2 & Few-shot & Three examples from templates T1, T2, T4, and then tested on templates T3, T5, T6; groups separated by visual similarity. \\
Iterative & Iterative & Sequential field-by-field extraction with tailored instructions; increases API calls but enables fine-grained control. \\
\hline
\end{tabular}
\end{table}

For the Cross-valid\_v2 experiment, templates were grouped by visual similarity: T1, T2, and T4 share tabular, highly structured layouts, whereas T3, T5, and T6 feature more dispersed, column-based designs. This design tests the model's ability to transfer extraction patterns to structurally different documents.

%-------------------------- Experiments ---------------------------
\section{Experimental Setup}
\label{se:experiments}

This section describes the experimental configuration, including the models evaluated, inference parameters tested, and evaluation metrics employed. The experimental design comprises two complementary analyses: a parameter sensitivity study examining how inference settings affect extraction quality and a comparative evaluation of prompting strategies across models. We evaluated two state-of-the-art large language models representing different architectural approaches to efficiency and scale:
\begin{itemize}
    \item \textbf{Gemini 1.5 Pro}~\cite{Reid2024gemini} is Google DeepMind's flagship model, built on a sparse Mixture-of-Experts (MoE) Transformer architecture that activates only a subset of parameters per inference, achieving computational efficiency without sacrificing quality. Its extended context window supports up to 1 million tokens with near-perfect information retrieval (>99.7\%).
    
    \item \textbf{Mistral-small}~\cite{Mistral2025small} is a dense 24-billion parameter model optimized through architectural innovations including Sliding Window Attention and Grouped-Query Attention, enabling a 128,000-token context window while maintaining efficiency suitable for high-throughput applications.
\end{itemize}

%Both models offered generous free API tiers at the time of experimentation, enabling nearly 500,000 API calls without prohibitive costs. 
These models were accessed through their respective cloud APIs without any fine-tuning, relying entirely on prompt engineering to adapt them to the extraction task. We investigated three key sampling parameters that control LLM text generation: Temperature, which controls output randomness by rescaling logits before softmax; Top-K, which limits candidate tokens to the $K$ most probable; and Top-P. nucleus sampling, which dynamically selects tokens whose cumulative probability exceeds the threshold $P$. To systematically evaluate parameter sensitivity, we designed a grid search over the parameter space within the recommended operating ranges for Gemini 1.5 Pro. Table~\ref{tab:parameter_grid} shows the 19 configurations tested: 18 combinations from the grid search plus one deterministic baseline using greedy decoding (Temperature = 0).

\begin{table}[htbp]
\centering
\caption{Inference parameter configurations evaluated. The grid search explores stochastic settings, while the deterministic baseline (ID 0) uses greedy decoding.}
\label{tab:parameter_grid}
\renewcommand{\arraystretch}{1.2}
\begin{tabular}{lccc}
\hline
\textbf{Configuration} & \textbf{Temperature} & \textbf{Top-P} & \textbf{Top-K} \\
\hline
Deterministic (ID 0) & 0.0 & -- & -- \\
\hline
Grid search & 1.0, 1.5, 2.0 & 0.5, 0.7, 0.95 & 32, 64 \\
(IDs 1--18) & \multicolumn{3}{c}{(all combinations: $3 \times 3 \times 2 = 18$)} \\
\hline
\end{tabular}
\end{table}

Due to resource constraints, the full parameter sensitivity analysis was conducted exclusively on Gemini 1.5 Pro, prioritizing depth of analysis on a single model. For the comparative prompting strategy evaluation, both models were run with the deterministic configuration to isolate the effect of prompt design from sampling variability.

The experimental design comprises two phases, summarized in Table~\ref{tab:experiment_summary}:
\begin{itemize}
    \item \textbf{Phase 1 - Parameter sensitivity analysis:} This phase evaluated the 19 parameter configurations on Gemini 1.5 Pro using two prompting strategies: zero-shot across all 2,400 invoices, and few-shot\_v1 on a subset of 800 invoices (templates T1 and T2). The objective was to quantify how inference parameters affect extraction accuracy and determine whether stochastic sampling offers advantages over deterministic decoding for this task.
    \item \textbf{Phase 2 - Prompting strategy comparison:} This phase compared all six prompting strategies using the deterministic configuration on both models. For few-shot strategies, results exclude templates used as examples to prevent artificially inflated performance: Few-shot\_v1 and Few-shot\_v2 exclude template T1 (used as the example), while the iterative strategy excludes templates T1 and T2 (used to develop field-specific instructions).
\end{itemize}

\begin{table}[htbp]
\centering
\caption{Summary of experimental configurations and API calls per model.}
\label{tab:experiment_summary}
\renewcommand{\arraystretch}{1.2}
\begin{tabular}{llrr}
\hline
\textbf{Phase} & \textbf{Configuration} & \textbf{Model} & \textbf{API Calls} \\
\hline
\multirow{2}{*}{Parameter analysis} & Zero-shot (19 configs $\times$ 2,400 invoices) & Gemini 1.5 Pro & 45,600 \\
 & Few-shot\_v1 (19 configs $\times$ 800 invoices) & Gemini 1.5 Pro & 15,200 \\
\hline
\multirow{4}{*}{Prompting strategies} & Non-iterative (5 strategies $\times$ 2,400 invoices) & Gemini 1.5 Pro & 12,000 \\
 & Few-shot\_v1 (1,600 remaining invoices) & Gemini 1.5 Pro & 1,600 \\
 & Iterative (83 calls/invoice $\times$ 2,400 invoices) & Gemini 1.5 Pro & 199,200 \\
 & Non-iterative (5 strategies $\times$ 2,400 invoices) & Mistral-small & 12,000 \\
 & Iterative (83 calls/invoice $\times$ 2,400 invoices) & Mistral-small & 199,200 \\
\hline
\multicolumn{2}{l}{\textbf{Total Gemini 1.5 Pro}} & & \textbf{273,600} \\
\multicolumn{2}{l}{\textbf{Total Mistral-small}} & & \textbf{211,200} \\
\hline
\end{tabular}
\end{table}

We evaluate extraction quality using standard information retrieval metrics: Precision, Recall, and F1-score, calculated at the field level. For each invoice, the model's JSON output is compared with the ground truth, and each field is classified as true positive (TP), false positive (FP), true negative (TN), or false negative (FN) based on string matching after normalization. Following the approach of~\cite{Sanchez2024bagofwords}, evaluation employs the template-specific label mapping described in Section~\ref{se:methods}, ensuring that models are only assessed on fields present in each document rather than being penalized for labels absent from a given template.

%-------------------------- Results ---------------------------
\section{Results}
\label{se:results}

This section presents the experimental results organized in three parts: First, we analyze the impact of inference parameters on extraction quality; then, we compare prompting strategies across both models; and finally, we examine per-label performance and the influence of document structure on extraction accuracy.

The parameter sensitivity analysis evaluated 19 configurations on Gemini 1.5 Pro to quantify the effects of temperature, Top-K, and Top-P on extraction performance. Table~\ref{tab:parameter_results} shows the aggregate metrics for each configuration.

\begin{table}[htbp]
\centering
\caption{Results of parameter tuning experiments on Gemini 1.5 Pro, showing Precision, Recall, and F1-Score for each configuration.}
\label{tab:parameter_results}
\renewcommand{\arraystretch}{1.2}
\begin{tabular}{ccc|ccc}
\hline
\textbf{Temp.} & \textbf{Top-K} & \textbf{Top-P} & \textbf{Precision} & \textbf{Recall} & \textbf{F1-Score} \\
\hline
0.0 & -- & -- & 80.87\% & 75.31\% & 77.99\% \\
1.0 & 32 & 0.50 & 81.40\% & 75.86\% & 78.54\% \\
1.0 & 32 & 0.70 & 81.33\% & 76.00\% & \textbf{78.57\%} \\
1.0 & 32 & 0.95 & 81.34\% & 75.59\% & 78.36\% \\
1.0 & 64 & 0.50 & 81.40\% & 75.89\% & 78.55\% \\
1.0 & 64 & 0.70 & 81.40\% & 75.84\% & 78.52\% \\
1.0 & 64 & 0.95 & 81.34\% & 75.63\% & 78.38\% \\
1.5 & 32 & 0.50 & 81.32\% & 75.94\% & 78.54\% \\
1.5 & 32 & 0.70 & 81.36\% & 75.96\% & \textbf{78.57\%} \\
1.5 & 32 & 0.95 & 81.11\% & 75.41\% & 78.16\% \\
1.5 & 64 & 0.50 & 81.31\% & 75.97\% & 78.55\% \\
1.5 & 64 & 0.70 & 81.29\% & 75.87\% & 78.49\% \\
1.5 & 64 & 0.95 & 81.16\% & 75.43\% & 78.19\% \\
2.0 & 32 & 0.50 & 81.40\% & 75.90\% & 78.56\% \\
2.0 & 32 & 0.70 & 81.27\% & 75.81\% & 78.45\% \\
2.0 & 32 & 0.95 & 81.09\% & 75.21\% & 78.04\% \\
2.0 & 64 & 0.50 & 81.34\% & 75.80\% & 78.47\% \\
2.0 & 64 & 0.70 & 81.33\% & 75.71\% & 78.42\% \\
2.0 & 64 & 0.95 & 81.08\% & 75.35\% & 78.11\% \\
\hline
\end{tabular}
\end{table}

The most notable observation is the remarkable stability of extraction performance across all parameter configurations. The variation ranges are minimal, with Precision varying by only 0.53 percentage points (from 80.87\% to 81.40\%), Recall by 0.79 points (75.21\% to 76.00\%), and F1-score by 0.58 points (77.99\% to 78.57\%). To contextualize these differences, with an average of approximately 60 extractable labels per invoice, a 1\% variation in precision corresponds to only 0.6 labels. Thus, the maximum observed variation means a difference of only 0.32 labels per invoice, which results in a negligible practical impact.

Despite the narrow range, a consistent pattern emerges: both extremes of determinism (Temperature = 0) and high randomness (Temperature = 2.0, Top-P = 0.95) yield suboptimal results. The best performance is achieved with moderate stochasticity, specifically Temperature = 1.0 or 1.5 with Top-P = 0.70, suggesting that a small degree of sampling diversity may help the model avoid suboptimal greedy choices without introducing excessive noise.

This finding has significant practical implications: for information extraction tasks requiring high fidelity, practitioners can safely use deterministic settings (Temperature = 0) to ensure reproducibility without significant performance degradation. The dominant factor influencing extraction accuracy is not the inference parameters but rather the inherent characteristics of each document template, as evidenced by the consistent performance hierarchy across templates regardless of parameter configuration.

The second experimental phase compared the six prompting strategies using deterministic inference (Temperature = 0) to isolate the effect of prompt design. Table~\ref{tab:gemini_results} presents the global metrics for Gemini 1.5 Pro.

\begin{table}[htbp]
\centering
\caption{Global metrics for Gemini 1.5 Pro across prompting strategies. Strategies marked with * exclude template T1 (used as example); IP excludes T1 and T2.}
\label{tab:gemini_results}
\renewcommand{\arraystretch}{1.2}
\begin{tabular}{lccc}
\hline
\textbf{Strategy} & \textbf{Precision} & \textbf{Recall} & \textbf{F1-score} \\
\hline
Zero-shot & 81.27\% & 75.71\% & 78.39\% \\
Few-shot\_v1* & 92.44\% & 87.88\% & 90.10\% \\
Few-shot\_v2* & 91.81\% & 87.30\% & 89.50\% \\
Cross-valid\_v1 & 96.18\% & 99.09\% & \textbf{97.61\%} \\
Cross-valid\_v2 & 95.24\% & 98.72\% & 96.95\% \\
Iterative (IP) & 94.45\% & 96.51\% & 95.47\% \\
\hline
\end{tabular}
\end{table}

The results reveal a clear performance hierarchy. Zero-shot extraction establishes a baseline with 78.39\% F1-score, confirming the inherent difficulty of the task without contextual examples. The introduction of in-context learning through few-shot strategies yields substantial improvements: both Few-shot\_v1 (annotated example) and Few-shot\_v2 (non-annotated example) exceed 89\% F1-score, representing 11+ percentage point gain over zero-shot.

Notably, the difference between annotated and non-annotated examples is negligible (90.10\% vs. 89.50\% F1-score), indicating that explicit label markers do not significantly improve the model's performance, making it possible to infer text-to-label correspondences from the input-output pair alone.

The cross-validation strategies achieve the highest performance. Cross-valid\_v1 employs sequential grouping, providing three examples from templates T1, T2, T3, and testing on templates T4, T5, T6; it reaches 97.61\% F1-score with Recall 99.09\%. Cross-valid\_v2 employs visual similarity grouping, providing examples from structurally similar templates T1, T2, T4, and testing on structurally different templates T3, T5, T6; it achieves 96.95\% F1-score. The comparable performance between these two approaches, one grouping sequentially, 
and the other by structural similarity, demonstrates that LLMs can generalize well regardless of whether in-context examples share exact visual characteristics with the target documents: similar structures also improve in-context learning.

The iterative extraction strategy (IP) achieves 95.47\% F1-score, which is lower than the cross-validation approaches. While decomposing extraction into field-specific queries provides fine-grained control, the additional complexity does not translate to superior aggregate performance, though it improves accuracy on specific problematic fields (discussed in Section~\ref{sec:per_label}). Table~\ref{tab:mistral_results} presents the corresponding results for Mistral-small.

\begin{table}[htbp]
\centering
\caption{Global metrics for Mistral-small across prompting strategies.}
\label{tab:mistral_results}
\renewcommand{\arraystretch}{1.2}
\begin{tabular}{lccc}
\hline
\textbf{Strategy} & \textbf{Precision} & \textbf{Recall} & \textbf{F1-Score} \\
\hline
Zero-shot & 76.59\% & 90.39\% & 82.92\% \\
Few-shot\_v1* & 90.31\% & 100.00\% & 94.91\% \\
Few-shot\_v2* & 89.55\% & 99.98\% & 94.48\% \\
Cross-valid\_v1 & 92.61\% & 99.88\% & \textbf{96.11\%} \\
Cross-valid\_v2 & 91.50\% & 99.75\% & 95.45\% \\
Iterative (IP) & 92.55\% & 98.14\% & 95.26\% \\
\hline
\end{tabular}
\end{table}

Mistral-small exhibits a similar pattern, with zero-shot performance presenting the lowest F1-score (82.92\%), few-shot strategies improving substantially (94.48--94.91\% F1-score), and cross-validation achieving the best results (96.11\% F1-score for Cross-valid\_v1). A distinctive characteristic of Mistral-small is its exceptionally high recall, reaching 100\% for Few-shot\_v1 and exceeding 99\% for most strategies, paired with consistently lower precision than Gemini 1.5 Pro. This pattern suggests higher hallucination rates in Mistral-small, as the model tends to output values for fields when they are not present in the document, generating false positives. In contrast, Gemini 1.5 Pro demonstrates greater restraint, more accurately identifying when a label is absent from the input, which may stem from its substantially larger parameter count and more sophisticated reasoning capabilities. This behavioral difference highlights a challenge of LLMs: smaller models may not abstain from extraction when evidence is insufficient, defaulting instead to producing plausible but incorrect outputs.

The performance gap between the two models is narrower than might be expected given their architectural differences. Despite Mistral-small being a considerably smaller and more resource-efficient model, it achieves results within 1--2 percentage points of Gemini 1.5 Pro across most strategies in terms of F1-score. However, the precision-recall trade-off reveals qualitatively different behaviors: Gemini optimizes for accuracy by avoiding false extractions, whereas Mistral prioritizes coverage at the cost of more hallucinated outputs. For applications where false positives carry significant cost (e.g., automated invoice processing without human review), Gemini's higher precision may be preferable; for tasks prioritizing completeness with downstream validation, Mistral's recall-oriented behavior may be the choice.

Both models demonstrate that the hierarchy of prompting strategies is consistent regardless of architecture: performance improves progressively from zero-shot, through single-example few-shot, to multi-example cross-validation. This consistency underscores that prompt quality dominates model choice and parameter tuning as the primary determinant of extraction accuracy for this case study.

%-----------New Condensed Subsection------------------------------------

\subsection{Per-Label Performance Assessment}
\label{sec:per_label}

To understand generalization capabilities at a finer granularity, we analyzed per-label precision using Cross-valid\_v2 results, which evaluate models on templates structurally different from the provided examples. Table~\ref{tab:label_precision} summarizes the distribution of labels across precision ranges.

\begin{table}[htbp]
\centering
\caption{Distribution of labels by precision range for Cross-valid\_v2 experiments.}
\label{tab:label_precision}
\renewcommand{\arraystretch}{1.2}
\begin{tabular}{lcc}
\hline
\textbf{Precision Range} & \textbf{Gemini 1.5 Pro} & \textbf{Mistral-small} \\
\hline
99--100\% & 61.9\% & 55.7\% \\
90--99\% & 20.6\% & 12.4\% \\
80--90\% & 7.2\% & 11.3\% \\
70--80\% & 2.1\% & 10.3\% \\
60--70\% & 2.1\% & 3.1\% \\
50--60\% & 2.1\% & 3.1\% \\
$<$50\% & 4.1\% & 4.1\% \\
\hline
$\geq$90\% (total) & \textbf{82.5\%} & \textbf{68.1\%} \\
\hline
\end{tabular}
\end{table}

Gemini demonstrates exceptional robustness, with 82.5\% of labels achieving precision above 90\% and 61.9\% with very high precision (99--100\%). Mistral shows lower performance, with 68.1\% of labels above 90\%, but still has a strong generalization capability. Both models struggle with a consistent set of problematic labels: C9 (website), CD (customer service phone), E8 (CNAE code), K2/K2d (energy rate), K4/K4d (power rate), M3/M3d (equipment rental price), and DD (distributor phone), characterized by low corpus frequency, semantic ambiguity, or placement in regions that degrade during PDF-to-Markdown conversion~\cite{Sanchez2024bagofwords}.

%\textbf{Structural influence on performance.} 
Template-level analysis reveals that document structure dominates extraction accuracy in zero-shot scenarios. Template T1 achieves the highest precision (86.91\% Gemini, 82.98\% Mistral), while T5 yields the lowest (77.67\% and 68.70\%, respectively). This disparity reflects preprocessing quality: T1's tabular layout preserves semantic relationships in Markdown, whereas T5's dense multi-column design degrades during conversion. Additionally, T5 contains several problematic fields (C9, E8, CD, M3), revealing structural challenges with content ambiguity.

\subsection{Comparison with Prior Work}

\begin{table}[htbp]
\centering
\caption{Comparison with classical ML baseline on IDSEM dataset. 
Classical ML was trained on templates T1--T6 and tested on T7; 
LLM cross-validation used examples from one template family (T1, T2, T4) 
and tested on structurally different templates (T3, T5, T6).}
\label{tab:comparison_baseline}
\renewcommand{\arraystretch}{1.2}
\begin{tabular}{llcc}
\hline
\textbf{Approach} & \textbf{Method} & \textbf{Seen Templates} & \textbf{Unseen Templates} \\
\hline
Sánchez \& Cuervo (2024) & SVM (RBF kernel) & 91.86\% & 67.20\% \\
This work (Gemini) & Cross-valid\_v2 & --- & 95.24\% \\
This work (Mistral) & Cross-valid\_v2 & --- & 91.50\% \\
\hline
\end{tabular}
\end{table}

The classical approach presented in~\cite{Sanchez2024bagofwords} achieved 91.86\% precision on templates seen during training but suffered severe degradation to 67.20\% when evaluated on template T7; see Table~\ref{tab:comparison_baseline}. This 24+ percentage points drop reveals significant overfitting to template-specific features. In contrast, LLMs require no template-specific training: the cross-validation experiments provided examples exclusively from one structural family (T1, T2, T4) and tested on a completely different family (T3, T5, T6). Under this evaluation, Gemini achieved 95.24\% precision, and Mistral achieved 91.50\%, both maintaining high accuracy on templates structurally dissimilar from the provided examples.

This comparison validates the hypothesis that foundational models can overcome the template-dependency limitations of classical approaches through their pre-trained semantic understanding and in-context learning capabilities, without requiring template-specific training data. Where classical ML memorizes layout-specific patterns that fail to transfer, LLMs abstract the extraction task itself, enabling robust generalization across different document structures.

Our experimental results yield four principal conclusions: First, prompt quality dominates extraction accuracy, as the 19+ percentage points improvement from zero-shot (78--83\% F1) to cross-validation few-shot (96--97\% F1) demonstrates that prompt engineering is the primary lever for maximizing performance; second, inference parameters have negligible impact, as the maximum F1-Score variation across 19 configurations (0.58 points) translates to fewer than 0.4 labels per invoice, allowing practitioners to use deterministic settings for reproducibility without performance loss; third, preprocessing quality critically affects results, since documents with clear tabular layouts (T1: 86.91\% zero-shot precision) significantly outperform complex multi-column designs that degrade during PDF-to-Markdown conversion (T5: 77.67\%); fourth, LLMs demonstrate robust cross-template generalization, maintaining 95--96\% precision on structurally dissimilar templates, whereas classical ML dropped 24+ points on unseen layouts, validating that LLMs generalize the extraction task rather than memorizing template-specific patterns.

%-------------------------- Conclusion ---------------------------
\section{Conclusion}
\label{se:conclusion}
This work shows that general-purpose LLMs achieve state-of-the-art performance on structured information extraction from semi-structured documents, surpassing classical machine learning approaches while exhibiting superior generalization to unseen document layouts. These approaches overcome the template-dependency of classical machine learning methods. With appropriately designed prompts, LLMs achieve superior extraction accuracy and generalization without task-specific fine-tuning on the IDSEM dataset.

The quality of prompts is the dominant factor determining extraction success, outweighing differences in model architecture and optimization of inference parameters. As LLMs continue to evolve, this finding suggests that prompt engineering will remain a critical competency for deploying general-purpose models in document understanding applications. 

While task-specific fine-tuning can internalize domain knowledge directly into model weights, reducing or eliminating the need for elaborate prompts, organizations leveraging off-the-shelf foundational models without custom training may still depend on sophisticated prompt design to bridge the gap between general capabilities and domain-specific requirements.

%This work presents some limitations. First, our parameter sensitivity analysis was conducted exclusively on Gemini 1.5 Pro due to resource constraints; replicating this analysis on Mistral-small would verify whether the negligible parameter impact generalizes to smaller architectures. 
This work presents some limitations. The validation on real-world invoices with authentic noise and variability would strengthen practical applicability, as the IDSEM dataset, although realistic, contains synthetic data. Another important limitation is the exclusive reliance on text-based processing. Converting PDF documents to Markdown discards visual and structural information that cannot be captured through text extraction alone. Vision-Language models, such as Donut~\cite{Kim2022donut}, and Layout-aware models, like LayoutLM~\cite{Xu2019layoutlm}, combine NLP with visual understanding, potentially recovering information lost in our text-only pipeline.

%Additionally, privacy considerations present deployment challenges. Sending sensitive business documents to cloud-based LLM providers introduces risk in regulated industries. For organizations with strict data governance requirements, on-premise deployment of fine-tuned smaller models offers a more secure alternative.

In future works, we will use hybrid prompting strategies to improve extraction efficiency with iterative field-specific precision, using routing prompts to identify fields that require detailed attention. We are also interested in multimodal approaches to process document images directly. We will also conduct a qualitative error analysis to distinguish format errors, semantic confusions, and hallucinations.

\bibliographystyle{unsrt} 
\bibliography{references}  

\end{document}